%% file: aaai2026.tex
\documentclass[letterpaper]{article} 
\usepackage{aaai2026}
\usepackage{times}  
\usepackage{helvet}  
\usepackage{courier}  
\usepackage[hyphens]{url}  
\usepackage{graphicx} 
\usepackage{natbib}  
\usepackage{caption} 
\usepackage{algorithm}
\usepackage{algorithmic}
\usepackage{amsmath}
\usepackage{booktabs}
\usepackage{multirow}
\usepackage{arydshln}
\usepackage{enumitem}
\usepackage{longtable}
\usepackage{xtab}
\usepackage{cuted}
\usepackage{pifont}
\usepackage{tabularx}

\usepackage{newfloat}
\usepackage{listings}
\DeclareCaptionStyle{ruled}{labelfont=normalfont,labelsep=colon,strut=off} 
\floatstyle{ruled}
\newfloat{listing}{tb}{lst}{}
\floatname{listing}{Listing}
\title{AgentMental: An Interactive Multi-Agent Framework for Explainable and Adaptive Mental Health Assessment}

\usepackage{bibentry}

\author{
Jinpeng Hu$^{1}$,
Ao Wang$^{1}$,
Qianqian Xie$^{3}$,
Hui Ma$^{1}$,
Zhuo Li$^{2}$,
Dan Guo$^{1}$
}
\affiliations{
    $^1$Hefei University of Technology \\
    $^2$The Chinese University of Hong Kong, Shenzhen \\
    $^{3}$Wuhan University \\
\{jinpenghu, huima, guodan\}@hfut.edu.cn, 2024170834@mail.hfut.edu.cn,\\
221019088@link.cuhk.edu.cn, xqq.sincere@gmail.com
}

\begin{document}

\nocopyright
\maketitle
\nocopyright

\begin{abstract}
Mental health assessment is crucial for early intervention and effective treatment, yet traditional clinician-based approaches are limited by the shortage of qualified professionals.
Recent advances in artificial intelligence have sparked growing interest in automated psychological assessment, yet most existing approaches are constrained by their reliance on static text analysis, limiting their ability to capture deeper and more informative insights that emerge through dynamic interaction and iterative questioning.
Therefore, in this paper, we propose a multi-agent framework for mental health evaluation that simulates clinical doctor-patient dialogues, with specialized agents assigned to questioning, adequacy evaluation, scoring, and updating.
We introduce an adaptive questioning mechanism in which an evaluation agent assesses the adequacy of user responses to determine the necessity of generating targeted follow-up queries to address ambiguity and missing information.
Additionally, we employ a tree-structured memory in which the root node encodes the user's basic information, while child nodes (e.g., topic and statement) organize key information according to distinct symptom categories and interaction turns.
This memory is dynamically updated throughout the interaction to reduce redundant questioning and further enhance the information extraction and contextual tracking capabilities.
Experimental results on the DAIC-WOZ dataset illustrate the effectiveness of our proposed method, which achieves better performance than existing approaches.
\end{abstract}

\section{Introduction}
Mental health issues significantly impact global well-being and represent a growing public health concern.
According to the World Health Organization (WHO), over 300 million people worldwide suffer from depression, and the prevalence of anxiety disorders reaches up to 4\% of the global population.
Mental health assessment plays a vital role in facilitating early detection and timely intervention for mental health disorders.
However, traditional mental health assessment methods rely on clinicians conducting structured interviews, such as PHQ-8, while they usually require professional personnel, which makes it difficult to access assessment services in resource-scarce regions.
Thus, many automated methods based on machine learning have been proposed to evaluate psychological states.

\begin{figure}[t]
  \centering
  \includegraphics[width=0.5\textwidth, trim=0 0 -20 0]{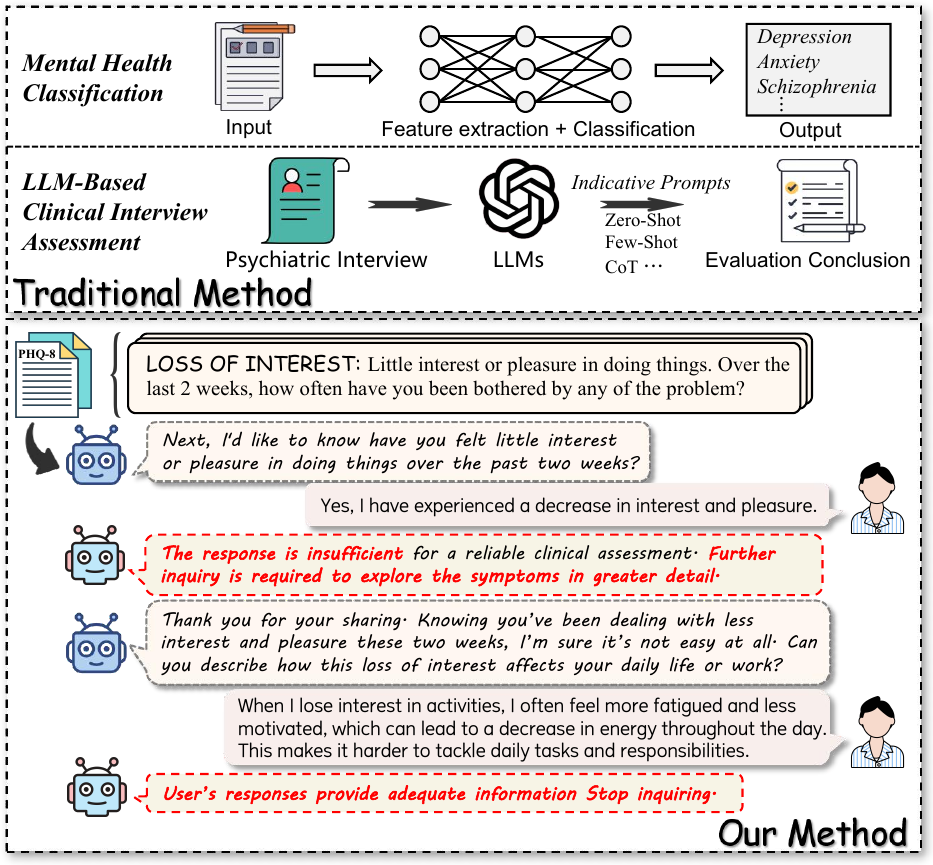}
  \caption{The comparison between our method and existing methods. }
  \label{fig:introduction}
\vspace{-1em}
  
\end{figure}

\begin{figure*}[t]
  \centering
  \includegraphics[width=\linewidth]{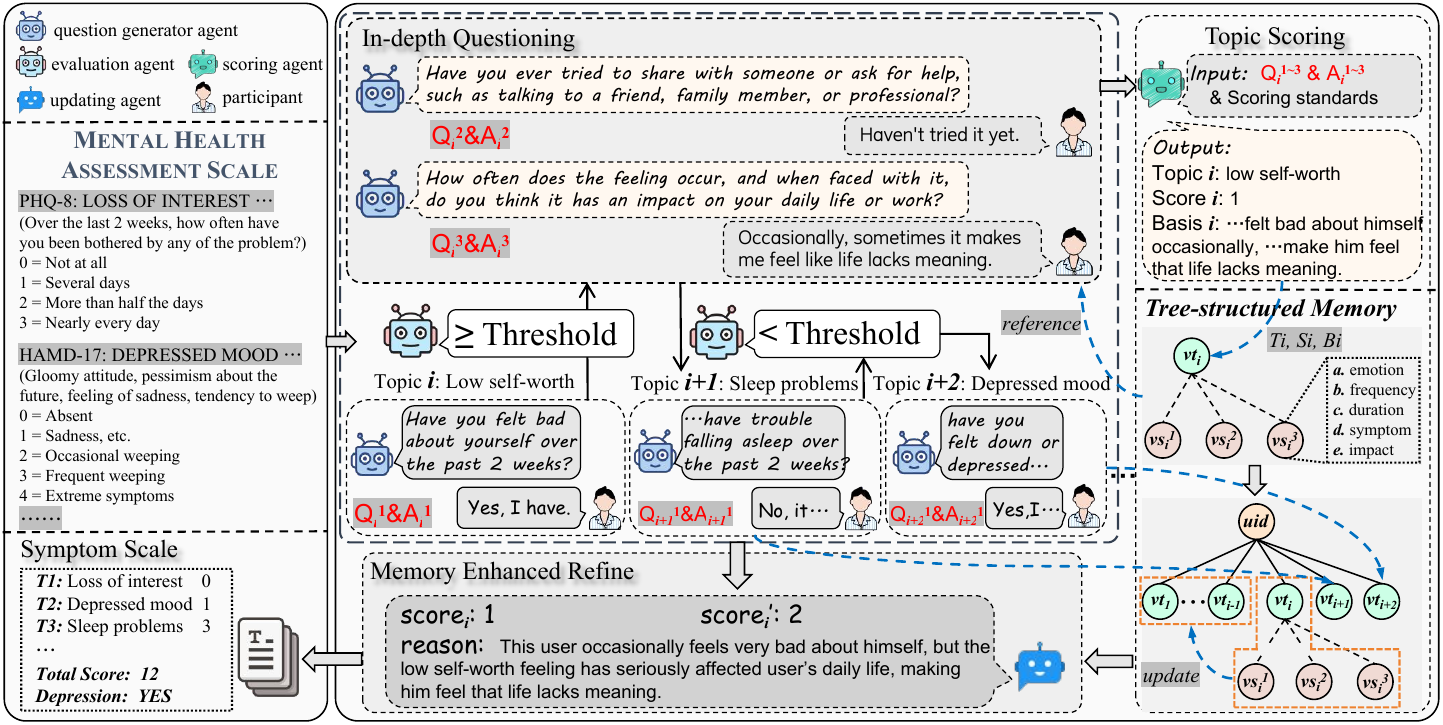}
  \caption{
  The overall framework of AgentMental for mental health assessment scales.
  }
  \label{fig:framework}
  \vspace{-1em}
\end{figure*}

Prior efforts have used text classification techniques in detecting a range of psychological disorders, including depression, suicidal ideation, bipolar disorder, and anxiety \cite{ji-etal-2022-mentalbert, 10681286,chiong2021textual,agarwal2024analysing}.
For example, \citet{10681286} introduced an emotion-aware contrastive network for mental disorder detection from social media posts.
\citet{agarwal2024analysing} designed multi-view architectures that separately model therapist and patient utterances to enhance automated depression estimation.
Recently, large language models (LLMs) have shown significant success in text comprehension and generation, leading to the development of LLM-based approaches for mental health \cite{yang2024mentallama, xu2024mental, li2024zero, xiao2024healme}.
For example, \citet{yang2024mentallama} constructed a mental health instruction dataset and trained MentaLLaMA for interpretable mental health analysis.
\citet{ohse2024gpt} investigated the performance of GPT-4 in identifying social anxiety from semi-structured clinical interview data.
\citet{chen2024depression} transformed the clinical interview into an expertise-inspired directed acyclic graph to improve depression detection.

Although these studies have achieved considerable progress, there are still two challenges.
First, most existing approaches focus on passive analysis of static data, limiting their ability to handle scenarios where user-provided information is incomplete or insufficient due to factors such as user resistance, or other contextual constraints.
Second, although standardized diagnostic scales are essential for depression assessment and clinical interpretability, the associated counseling process often involves heterogeneous and multi-topic data, making it challenging for models to derive a holistic and coherent understanding of the user's mental state.
Recent developments in LLM–based role-playing have demonstrated robust performance across diverse tasks, providing a technological foundation for effectively simulating interactive dialogues and thereby enabling the acquisition of more comprehensive information \cite{yu-etal-2025-beyond, jiang-etal-2024-personallm}.

Therefore, in this paper, we propose a multi-agent-based interactive framework to simulate clinical interview scenarios for mental health assessment (AgentMental), wherein dedicated agents are responsible for distinct roles such as question generation, response evaluation, score assignment, and updating.
Prior to score assignment, we use an adaptive questioning mechanism to alleviate the issue of insufficient information in user responses, wherein an evaluation agent assesses the completeness of user responses to determine whether the question generation agent should produce targeted follow-up queries aimed at eliciting more comprehensive information.
Furthermore, given the contextual dependencies inherent in scale-based assessments, we introduce a tree-based memory to systematically record critical information, in which the root node contains fundamental user information, and topic nodes store item-level assessment scores and dialogue summaries, and statement nodes preserve turn-level evidence, such as symptom frequency, duration, etc.
By dynamically updating this memory structure throughout the interactive process, the proposed method effectively reduces redundant questioning, facilitates coherent information flow across different topics, and consequently enhances the overall accuracy of the psychological assessment.
In addition, we conduct comprehensive evaluations on the DAIC-WOZ dataset, and the results demonstrate that AgentMental consistently outperforms existing studies.

\section{Methodology}

In this section, we introduce the details of the proposed multi-agent-based framework designed for automated mental health assessment, including tree-structured memory, adaptive question generation, and memory-augmented scoring.
The overall framework is shown in Figure \ref{fig:framework}.

\subsection{Overall Workflow}

\paragraph{Motivation.}
Based on discussions with psychologists, real-world psychological scale evaluations typically involve clinicians asking patients topic-related questions and subsequently assigning a rating score based on their responses \cite{sommers2015clinical}. 
This process allows psychologists to assess various aspects of a patient's mental state in a structured yet flexible manner.
Following this, to simulate this real-world evaluation process, and for each given psychological scale, we generate an initial question for each topic to ensure comprehensive coverage of key assessment criteria. 
Then, based on the user response, we dynamically determine whether additional follow-up questions are needed for the current topic, facilitating a more in-depth evaluation of the individual's mental state.
Upon completion of the conversation, the system adaptively updates the scores along with explanatory reasoning and generates a psychological report that includes a dialogue summary and personalized recommendations.

\paragraph{Preparation.}

AgentMental incorporates four core agents, each fulfilling distinct roles within the system, including question generator agent $AG_q$, evaluation agent $AG_{ev}$, scoring agent $AG_s$, updating agent $AG_u$.
The question generator agent is responsible for generating core assessment questions to ensure the structured progression of the dialogue.
The evaluation agent performs necessity analysis and demand judgment, evaluating the completeness and depth of user responses to determine whether further questioning is required. 
The scoring agent focuses on analyzing user responses and generating quantitative scores for symptom severity based on clinical standards. 
The updating agent is responsible for information integration and semantic compression, as well as adjusting assessment scores based on the aggregated interaction history.

\subsection{Tree-Structured Memory}

Given the interdependencies between scale topics, we use a tree-structured memory that evolves throughout the conversation to record critical information, enhance information retention and reduce redundancy of questions.
The memory consists of three types of nodes: a user node representing basic user attributes (e.g., occupation, sex, age and so on), topic nodes $vt$ corresponding to individual assessment topics (e.g., score, summary and so on), and statement nodes $vs$ encoding salient information extracted from user responses (e.g., emotions, frequency, duration, symptoms, impact). 
At each dialogue turn, several key dimensions of information, including emotion, frequency, duration, symptoms, and impact, are extracted by a LLM and a corresponding statement node is constructed to retain this information.
Let $i$ denote the index of the current topic.
Upon completion of a $i$-th topic $T_i$, the following operations are performed:
\begin{itemize}
    \item Instantiates a topic node $vt_i$ recording the topic-specific assessment score $S_i$ and behavioral summary $B_i$ generated by the agent $AG_s$.
    \item Updates summary sections of previously instantiated topic nodes ${vt_1, \ldots, vt_{i-1}}$ to incorporate new relevant information.
    \item Establishes edges to all statement nodes associated with topic $T_i$.
\end{itemize}
Besides, each topic node is also connected to the user node, thereby enabling context-aware filtering of inappropriate questions.
For instance, querying a primary or secondary school student about work-related stress would be considered unsuitable, as individuals in this demographic are typically not engaged in formal employment.

\subsection{Adaptive Question Generation}
\label{question_generation}
We first extract the communication topic for each scale, such as PHQ-8 which contains 8 topics.
Then, the question generator agent $AG_{q}$ generates topic-specific questions to simulate interactions between psychologists and help-seekers, aiming to guide the user to give responses that can reflect their status on the given topic.
Besides, in clinical settings, it is common for patients to face difficulties in clearly expressing their thoughts and emotions due to cognitive impairments, resistance, or communication barriers.
For instance, a patient with severe depression may respond with minimal or vague answers, such as simply saying ``I do have insomnia'' when asked about their symptoms, without providing key details such as the frequency or duration of insomnia.
Therefore, more specific inquiries are needed to elicit more comprehensive information from the patient, such as ``Does this happen frequently?'' or ``How long does it take you to fall asleep?''.
To replicate this dynamic interaction, we incorporate a context-aware follow-up mechanism, which can detect vague, ambiguous, or incomplete responses and generate appropriate clarification questions to guide the conversation.
In detail, we use the evaluation agent $AG_{ev}$ to systematically assess patient responses in terms of information completeness and symptom severity.
It produces a necessity score on a scale from 0 to 2, representing the degree to which further questioning is warranted for the current topic.
If the agent determines that the current information is insufficient for an accurate assessment (i.e., the score exceeds a threshold $\theta$), $AG_{q}$ further dynamically formulates targeted follow-up questions to encourage the patient to provide additional details.
For follow-up questions, we incorporate several additional constraints:
\begin{itemize}
    \setlength{\topsep}{0pt}
    \setlength{\itemsep}{0pt}
    \setlength{\parsep}{0pt}
    \setlength{\parskip}{0pt}
    \item The in-depth questions require actively guiding the patient to elaborate on the severity, frequency, duration, and impact of their symptoms.
    \item The agent is encouraged to generate easy-to-answer questions, aiming to reduce the user's cognitive burden and psychological resistance.
\end{itemize}

Let $j$ represent the index of the follow-up question within $i$-th topic.
Then in-depth question generation process can be formally defined as:
\begin{equation}
    Q_{i}^{j+1} = AG_{q} (T_{i}, H_{i}^{j}, M_{i}^{j}),
    \label{equal_question_generation}
\end{equation}
$H_{i}^{j}$ represents the contextual history of preceding question-answer pairs.
The memory state $M_{i}^{j}$ maintains a structured record of the accumulated dialogue history up to the $(i{-}1)$-th topic, encompassing topic-level summaries and scores from $T_1$ to $T_{i-1}$, as well as salient information extracted from all prior conversational turns within $i$-th topic, spanning from $(Q_{i}^{1}, A_{i}^{1})$ to $(Q_{i}^{j}, A_{i}^{j})$.

Once sufficient information is collected or the predefined upper limit of follow-up questions $d$ is reached, the system automatically transitions from the in-depth questioning phase to the next topic, ensuring an efficient and structured evaluation.
Therefore, the question generation process can be formulated as:
\begin{equation}\label{fitting process}
    Q_{next}=\begin{cases}
    Q_{i}^{j+1}, &  AG_{ev}(Q_i^{1 \sim j},A_i^{1 \sim j})>\theta \land j < d \\
                       Q_{i+1}^{1}, & \text{otherwise}
                        \end{cases}
\end{equation}

\subsection{Memory-Augmented Scoring}
\label{memory}
After the complete conversation about a topic, the scoring agent is applied to assess the patient state.
\begin{equation}
    [S_{i}, B_{i}] = AG_{s} (Q_i^{1 \sim N_{i}}, A_i^{1 \sim N_{i}}, SR)
    \label{rating_scaling}
\end{equation}
where $S_{i}$ is the score and $B_{i}$ denotes the summary and supporting evidence, and $N_{i}$ is the number of questions for topic $T_{i}$. $SR$ refers to the scale rating standards.
For instance, regarding the ``depressed mood'', the scoring agent assigns a score of $1$, with the summary and supporting evidence as: ``The user described feeling down in the last two weeks, but only occasionally.''
The procedure for scoring and memory updating is presented in Algorithm~\ref{alg:topic_evaluation}.

After all topics in the scale are evaluated with their respective scores and memory, a comprehensive update is performed to account for inter-topic correlations.
The updating process can be formulated as:
\begin{equation}
    [S^{\prime}, R] = AG_{u}(H, M_{topic}),
\end{equation}
where $H$ denotes the overall dialogue context, including all questions and responses, $M_{topic}$ indicates aggregated information in topic nodes.
$S^{\prime}$ represents the final assessment scores assigned to each topic and $R$ represents the reasoning trace generated during the update process, as well as a comprehensive summary and professional suggestions for the current participant, which contributes to model explainability.
To derive the overall outcome, scores from all topics in $S^{\prime}$ are aggregated into a total score, which is then mapped to a diagnostic category based on the corresponding psychological scale's criteria.

\input{tab/algorithm}

\section{Experiments}

\subsection{Experimental Setting}
\noindent \textbf{Dataset.}
We conduct experiments on the Distress Analysis Interview Corpus-Wizard of Oz (DAIC-WOZ) \cite{gratch-etal-2014-distress}, a widely adopted English-language dataset for clinical depression detection. 
This corpus includes interviews with 189 participants, captured through transcripts, audio recordings, and videos. 
In this work, we focus exclusively on the text modality derived from the interview transcripts.
Following each interview, participants complete the PHQ-8, which evaluates eight depression-related symptoms.
Participants with a PHQ-8 score of 10 or higher are classified as depressed, while those scoring below 10 are labeled as control.
Following prior studies \cite{chen-etal-2024-depression}, we adhere to the fixed data split of DAIC-WOZ and use the development set for evaluation.
See Appendix A for details of the dataset.

\noindent \textbf{Baselines.}
To explore the effectiveness of our proposed method, we use the following prompt-based studies as our baselines, including Zero-Shot, Few-Shot, Chain-of-Thought (CoT) \cite{wei2022chain}, and Chain-of-Logic \cite{zhao-etal-2024-enhancing-zero}. 
In addition to prompt-based methods, we extended our comparative analysis to include several basic multi-agent-based approaches, which include MDAgents \cite{Kim2024MDAgentsAA}, Debate \cite{tran2025multiagentcollaborationmechanismssurvey}, and Majority Voting \cite{wang2023selfconsistency}.
See Appendix D for a detailed introduction.
For the aforementioned LLM-based methods, we implement two distinct experimental settings for effective comparison.
In the first setting, the model performs binary classification using the full interview transcript as input, denoted as $bc$.
In the second setting, the model predicts item-wise scores based on the interview text and relevant psychological scale topics. These item-level predictions are then aggregated into a total score, which is mapped to a categorical label according to the criteria defined by the scale, denoted as $sl$.
There are also some task-specific learning-based approaches, including Multi-View \cite{agarwal2024analysing}, and SEGA \cite{chen2024depression}. 

\input{tab/daic}

\noindent \textbf{Evaluation.}
In our experiments, we leverage transcribed text from the DAIC-WOZ dataset as prior knowledge and prompt LLMs to simulate individuals with or without mental disorders.
Subsequently, our proposed interactive framework engages with the simulated individuals, implementing multi-turn dialogues that reflect realistic mental health scenarios.
In validating the efficacy of prediction scores, we use mean absolute error (MAE) to evaluate precision between the predicted scores and the ground truth.
Additionally, based on the classification outputs, model performance is evaluated using Cohen’s Kappa coefficient (Kappa), the F1 scores for the Control and Depressed classes (denoted as F1[C] and F1[D]), and the Macro F1 score.
See Appendix B.1 for a detailed introduction for these metrics.

\input{tab/daic-item}

\noindent \textbf{Implementation Details.}
We adopt the AutoGen framework \cite{wu2024autogen} to implement our multi-agent system\footnote{https://github.com/microsoft/autogen}.
We employ the Qwen2.5 series (e.g., Qwen2.5-14B-Instruct and Qwen2.5-72B-Instruct) models as the core foundation models for comparative experiments, with the temperature parameter set to $0$ \cite{yang2024qwen2}.
Besides, we use Deepseek-R1 (i.e., Deepseek-R1-Distill-Qwen-32B) to simulate users \cite{guo2025deepseek}.
The experiments are conducted on an NVIDIA A6000 GPU platform, utilizing the vLLM framework \cite{kwon2023efficient} for acceleration.
The hyperparameters $d$ and $\theta$ are set to 3 and 1, respectively.

\subsection{Overall Performance}

To explore the effectiveness of our proposed method, we compare it with the aforementioned studies, with the results reported in Table~\ref{daic}.
There are several observations.
First, given the same model architecture, models with more parameters tend to have better performance, as demonstrated by the comparison between Qwen2.5-14B-Instruct and Qwen2.5-72B-Instruct.
This observation aligns with prior findings that larger models tend to demonstrate enhanced capabilities in language understanding and reasoning \cite{team2024qwen2}.
Second, methods evaluated under the $sl$ configuration consistently demonstrate superior performance compared to those under the $bc$ configuration.
This performance disparity can be attributed to the structured guidance offered by the psychological scale in the $sl$ setting, which enhances the extraction of critical information and reduces semantic ambiguity present in long-form psychological interviews.
Third, models augmented with prompt engineering outperform their non-augmented counterparts.
This improvement may stem from the model's ability to better interpret user input, infer underlying mental states, and produce more accurate and reliable assessment results.
Fourth, multi-agent-based approaches (i.e., Majority Voting, Debate, MDAgents and AgentMental) further exhibit superior performance than most prompt-based methods, which underscores the benefits of collaborative agent coordination.
AgentMental achieves the best performance in most evaluation metrics, confirming the effectiveness of integrating an in-depth questioning mechanism with a tree-based memory structure.
The in-depth questioning mechanism is capable of generating easy-to-answer yet informative questions, guiding users to provide more valuable and comprehensive information for better understanding of their mental state.
Meanwhile, the memory module helps reduce information redundancy and strengthens the coherence and interdependence of the collected information.

\subsection{Item-level Fine-Grained Analysis}

To further validate the framework’s capability to capture subtle differences in psychological states, we conducted a fine-grained analysis of 8 topics within the PHQ-8 scale, with the results shown in Table~\ref{tab:daic-item}.
First, AgentMental demonstrates superior performance across the majority of scale topics compared to the CoT and MDAgents, demonstrating its enhanced effectiveness.
Second, substantial performance gains are observed for symptoms such as ``Fatigue or Low Energy'' and ``Sleep Problems'', which often involve observable physiological or behavioral indicators.
These improvements suggest that our method benefits from its interactive design, where the ability to pose follow-up questions enables more precise and contextually grounded assessment.
Third, items characterized by high subjectivity and contextual dependency, such as ``psychomotor changes'', remain challenging to assess reliably within automated systems.
This difficulty likely stems from their reliance on subtle linguistic cues, including emotional tendencies throughout the interaction and signs of cognitive dissonance, which place higher demands on the model's reasoning and understanding capabilities.

\input{tab/ablation}

\subsection{Ablation Study}
\label{ablation_study_}
To explore the effectiveness of different modules in our proposed method, we conduct an ablation study comparing the full model to three variants: \uppercase\expandafter{\romannumeral1} without in-depth questioning mechanism, \uppercase\expandafter{\romannumeral2} without memory to store information, \uppercase\expandafter{\romannumeral3} without both of them.
The results are reported in Table~\ref{tab:ablation}.
It is observed that removing any component from our proposed method results in a noticeable decline in performance, highlighting the contribution of each component in enhancing the framework's ability to assess mental health.
Furthermore, the integration of in-depth questioning and memory allows the full model to outperform the variants, validating our design choice of combining these modules into a multi-agent framework.
This may be attributed to the complementary benefits of the two components: in-depth questioning facilitates the elicitation of richer user information, while the memory enhances the model’s ability to capture inter-topic relationships and retain relevant contextual details.

\subsection{GPT/Human Evaluation}

\begin{figure}[t]
    \centering
    \includegraphics[width=\columnwidth]{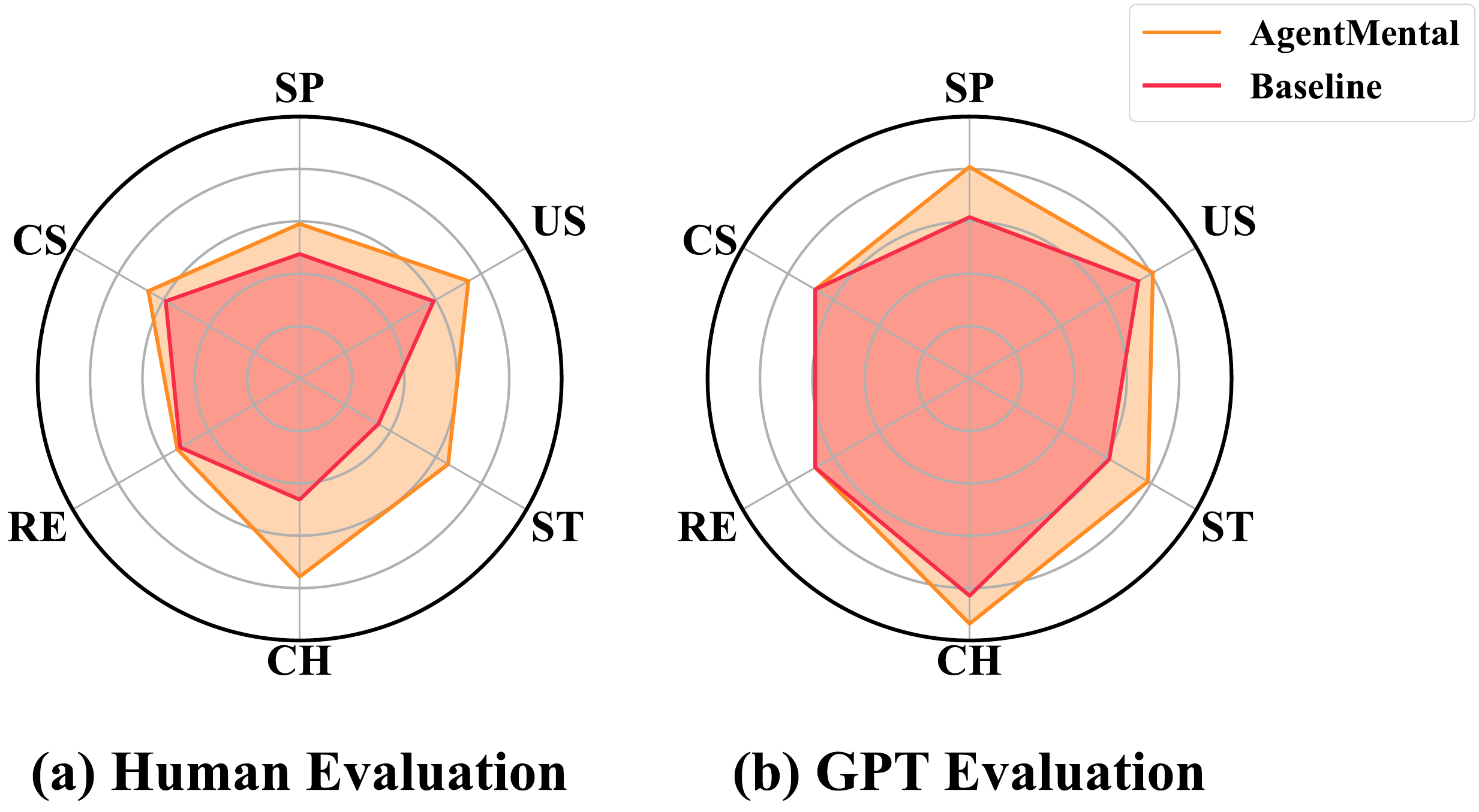}
    \caption{Comparison of our method with the baseline method through automatic and human evaluations, where US, SP, CS, RE, CH, ST refer to Understanding, Sympathy, Consistency, Realism, Coherence, and Satisfaction.}
    \label{fig:evaluation2}
    \vspace{-1em}
\end{figure}

To evaluate clinical utility and interaction quality, we conduct a double-blind evaluation experiment involving human and GPT assessments across 5 participants cases with varying symptom severities, comparing its performance against baseline methods (i.e., model \uppercase\expandafter{\romannumeral3} in \textbf{Ablation Study}).
Each case comprises the full dialogue text, assessment scale scores, a summary of the participant’s symptoms, and corresponding suggestions.
We adopt the multi-dimensional empathy framework and use Understanding and Sympathy to assess perceived empathy during psychological interviews \cite{xu-jiang-2024-multi}.
In addition, we follow \citet{yang-etal-2024-psychogat} to measure simulation effectiveness and user experience using four metrics: Consistency, Realism, Coherence, and Satisfaction.
Details for these metrics can be found in Appendix B.2.
\begin{itemize}
    \item For human evaluation, we employ five graduate students independently to review each case, and the average value is taken as the final result to reduce individual bias and ensure reliability.
    \item For GPT-4o automated evaluation, each case is scored independently three times, and the average value is taken as the final result.
\end{itemize}
The evaluation results are presented in Figure~\ref{fig:evaluation2}.
It is observed that our method outperforms the baseline across most aforementioned metrics, further demonstrating the effectiveness of our design for psychological evaluation.
Besides, the model demonstrates significantly greater improvements in Coherence, and Satisfaction relative to other measures.
This improvement may be attributed to the interactive design of AgentMental, which promotes more engaging, contextually coherent, and user-centered dialogue. 
Such an interaction paradigm further enhances users’ perceived satisfaction.

\begin{figure*}[t]
  \centering
  \includegraphics[width=\linewidth]{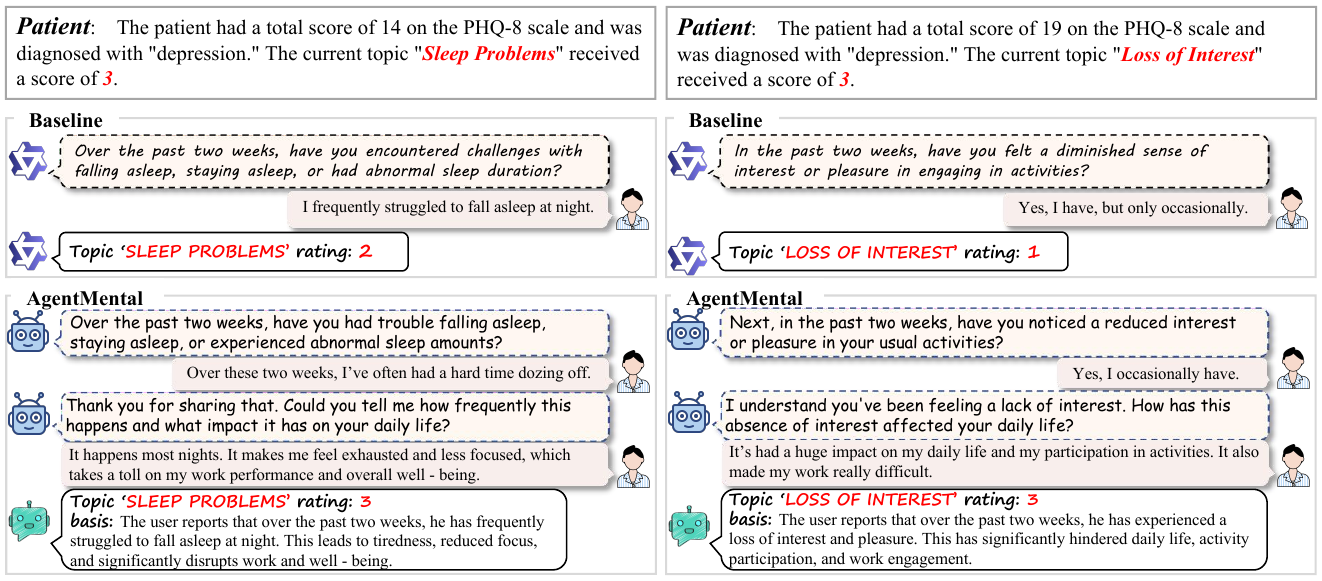}
  \caption{
  Examples of PHQ-8 topics ``SLEEP PROBLEMS'' and ``LOSS OF INTEREST'' from Baseline and AgentMental.
  }
  \label{fig:case}
  \vspace{-1em}
\end{figure*}

\subsection{Case Study}
To further investigate the effectiveness of our method, we conduct a qualitative analysis of two representative cases.
As shown in Figure~\ref{fig:case}, we present a comparison of the interaction processes and rating outcomes between our proposed method and the baseline model (i.e., Qwen2.5-72B-Instruct) in evaluating the topic of ``Sleep Problems'' and ``Loss of Interest'' on the PHQ-8.
In these cases, the baseline employs a single-question approach that captures only surface-level responses, whereas AgentMental utilizes multi-turn interactions to elicit deeper and more informative user input.
Triggered by the evaluation agent, follow-up inquiries aim to explore the underlying causes of symptoms and uncover critical details, thereby refining the assessment and demonstrating the framework’s capacity for dynamic adaptation to user-provided input.
Through the collection of the key information, e.g.,(``leaves him tired and less focused···work and well-being'' in the left example and ``significantly impacting daily life···activities and work'' in the right case), AgentMental can give a more precise score.

\section{Related Work}

\subsection{Multi-Agent Systems}

Multi-agent systems \cite{dorri2018multi} involve multiple autonomous agents that collaborate or compete to solve complex problems, adapting dynamically to changing environments \cite{li2023camel,chen2024agentverse, chan2024chateval,ijcai2024p890,wang-etal-2024-rethinking-bounds}.
Recently, multi-agent systems have been increasingly applied in the medical domain to enhance assessment and intervention by simulating interactions between patients and virtual agents \cite{fan-etal-2025-ai,almansoori2025self}.
For example, the MDAgents \cite{Kim2024MDAgentsAA} leverages a dynamic team of collaborative agents to develop highly personalized, comprehensive treatment plans for patients with multiple chronic conditions.
Similarly, the PsychoGAT framework \cite{yang-etal-2024-psychogat} employs agents to transform traditional questionnaires into interactive, game-based assessments, thereby boosting engagement and accuracy.
Furthermore, the INCHARACTER framework \cite{wang2023incharacter} assesses the personality consistency of role-playing agents through structured interviews, demonstrating the potential of MAS in measuring psychological traits.

\subsection{LLM for Mental Health}

In recent years, Large Language Models (LLMs) have shown great potential in natural language processing, leveraging extensive datasets and advanced architectures to comprehend and generate human-like text.
Models such as GPT-4 \cite{achiam2023gpt}, and LLaMA \cite{Touvron2023LLaMAOA} exhibit near-human capabilities in diverse tasks \cite{chowdhery2023palm}, including text generation, text classification, sentiment analysis, and contextual understanding.
In the realm of mental health, LLMs also offer significant potential in powering mental healthcare domain, such as diaglogue generation, depression and anxiety detection \cite{yang2024mentallama,xu2024mental}.
For example, MeChat \cite{qiu-etal-2024-smile}, which simulates psychotherapy dialogues to provide emotional support.
Furthermore, the LLAMADRS framework \cite{kebe2025llamadrs} utilizes LLMs to analyze clinical interview transcription data, achieving depression symptom scoring close to human assessment.
\citet{chen2024depression} transforms clinical interviews into a directed acyclic graph and improve it with principle-guided data using LLMs.
Although these approaches have brought significant improvements, they often lack explicit explainability and fail to guide patients to gradually disclose more valuable information during evaluation.
Compared to these methods, our method offers an alternative solution to introduce an in-depth questioning mechanism to actively elicit richer information from patients, mitigating cognitive barriers or resistance, while simultaneously providing a transparent process to enhance the explainability of the evaluation.

\section{Conclusion}

In this paper, we propose a multi-agent framework designed to simulate clinical doctor–patient dialogues for psychological evaluation.
Specifically, we integrate an adaptive in-depth questioning mechanism designed to elicit richer user information, dynamically adjusting its queries based on the adequacy evaluation of user responses.
Besides, we introduce a dynamic tree-based memory to structurally record critical information from previous dialogue turns, which is used to reduce question duplication and further improve performance.
Experimental results on the DAIC-WOZ dataset demonstrate that our proposed method outperforms existing baselines, showcasing its significant potential for advancing automated mental health assessment.


\bibliography{aaai2026}

\end{document}

%% file: tab/algorithm.tex
\begin{algorithm}[tb]
\caption{Topic Scoring}
\label{alg:topic_evaluation}
\textbf{Input}: Scale topics $T$, Scale rating standards $SR$, User ID $uid$\\
\textbf{Output}: Memory $M$
\begin{algorithmic}[1]

\STATE $M \leftarrow Initialize Memoy(uid)$
\STATE $d \leftarrow 3$, $i \leftarrow 1$ \COMMENT{Initialize time step}

\FOR{each theme $T_i$ in $T$}
  \STATE Initialize $j \leftarrow 0$
  
  \WHILE{$j < d$}
    \STATE $Q_{i}^{j+1} \leftarrow AG_{q}(T_{i}, H_{i}^{j}, M_{i}^{j})$
    \STATE $A_i^{j+1} \leftarrow \text{User}(Q_{i}^{j+1})$
    \STATE $\text{\{emotion, frequency, etc\}} \leftarrow LLM(A_i^{j+1})$
    \STATE $M.\text{add}(vs_i^{j+1})$ \COMMENT{Create the statement node}
  
    \STATE $\text{necessity} \leftarrow AG_{ev}(T_i, Q_i^{1 \sim {j+1}}, A_i^{1 \sim {j+1}})$
    \IF{$\text{necessity} < \theta$}
      \STATE \textbf{break}
    \ENDIF
    \STATE $j \leftarrow j + 1$

  \ENDWHILE
  
  \STATE $(S_i, B_i) \leftarrow AG_{s}(Q_i^{1 \sim j}, A_i^{1 \sim j}, SR)$
  \STATE $M.\text{add}(vt_i)\{T_i, S_i, B_i\}$

  \STATE $M.\text{update}(vt_1, vt_2,...,vt_{i-1})$
  
  \STATE $i \leftarrow i + 1$
\ENDFOR

\STATE \textbf{return} $M$

\end{algorithmic}

\end{algorithm}

%% file: tab/daic.tex
\begin{table*}[t]
  \centering
  \begin{tabular}{l|lccccc}

  \toprule

  \textbf{LLM} & \textbf{Methods} & \textbf{MAE}$\downarrow$ & \textbf{Kappa}$\uparrow$ & \textbf{F1[C]}$\uparrow$ & \textbf{F1[D]}$\uparrow$ & \textbf{Macro F1}$\uparrow$ \\

  \midrule
    \multirow{3}*{-}  
                                 & Multi-View              & - & - & - & - & 77.0 \\
                                 & SEGA                    & - & - & 88.4 & 81.5 & 84.9 \\
                                 & SEGA++                  & - & - & 90.9 & 84.6 & 87.8 \\
    \midrule

    \multirow{8}*{\rotatebox{90}{\parbox{3cm}{\centering Qwen2.5-14b}}} 

     & Zero-Shot ($bc$/$sl$)          & - / 3.314 & 51.2 / 56.8 & 81.8 / 88.5 & 69.2 / 66.7 & 75.5 / 77.6  \\
     & 5-Shot ($bc$/$sl$)              & - / 3.086 & 63.4 / 66.3 & 86.4 / 89.8 & 76.9 / 76.2 & 81.6 / 83.0  \\
     & CoT ($bc$/$sl$)                & - / 3.143 & 56.5 / 64.8 & 84.4 / 90.2 & 72.0 / 73.7 & 78.2 / 81.9  \\
     & Chain-of-Logic ($bc$/$sl$)     & - / 3.000 & 62.0 / 66.3 & 87.0 / 89.8 & 75.0 / 76.2 & 81.0 / 83.0  \\

     & Majority Voting ($bc$/$sl$)    & - / 2.857 & 56.5 / 62.0 & 84.4 / 87.0 & 72.0 / 75.0 & 78.2 / 81.0  \\
     & Debate ($bc$/$sl$)             & - / 2.857 & 51.2 / 67.7 & 81.8 / 89.4 & 69.2 / 78.3 & 75.5 / 83.8  \\
     & MDAgents ($bc$/$sl$)           & - / 2.629 & 62.0 / 67.7 & 87.0 / 89.4 & 75.0 / 78.3 & 81.0 / 83.8  \\

     & AgentMental                & 2.543 & 72.4 & 92.0 & 80.0 & 86.0  \\

  \midrule
\multirow{8}*{\rotatebox{90}{\parbox{3cm}{\centering Qwen2.5-72b}}}    
  
   & Zero-Shot ($bc$/$sl$)        & - / 3.286 & 53.0 / 56.8 & 81.0 / 88.5 & 71.4 / 66.7 & 76.2 / 77.6 \\
   & 5-Shot ($bc$/$sl$)           & - / 3.029 & 63.4 / 66.3 & 86.4 / 89.8 & 76.9 / 76.2 & 81.6 / 83.0 \\
   & CoT ($bc$/$sl$)              & - / 3.029 & 58.1 / 64.8 & 83.7 / 90.2 & 74.1 / 73.7 & 78.9 / 81.9 \\
   & Chain-of-Logic ($bc$/$sl$)   & - / 3.114 & 63.4 / 64.8 & 86.4 / 90.2 & 76.9 / 73.7 & 81.6 / 81.9  \\

   & Majority Voting ($bc$/$sl$)  & - / 2.771 & 63.4 / 67.7 & 86.4 / 89.4 & 76.9 / 78.3 & 81.6 / 83.8 \\
   & Debate ($bc$/$sl$)           & - / 2.800 & 54.7 / 68.9 & 80.0 / 88.9 & 73.3 / 80.0 & 76.7 / 84.4 \\
   & MDAgents ($bc$/$sl$)         & - / 2.800 & 63.4 / 73.6 & 86.4 / 91.7 & 76.9 / 81.8 & 81.6 / 86.7 \\

   & AgentMental & \textbf{2.514} & \textbf{79.8} & \textbf{$\text{93.9}^{*}$} & \textbf{$\text{85.7}^{*}$}  & \textbf{$\text{89.8}^{*}$}  \\

  \bottomrule
  \end{tabular}
  \caption{
    Comparison of the evaluation results of our method with baseline methods on the dataset DAIC-WOZ. \textbf{Bold} indicates best performance. $bc$ represents binary classification, and $sl$ represents item-level predictions. Results with $*$ are better than the baseline ($p < 0.05$) based on a one-tailed unpaired t-test.
  }
  \label{daic}
  \vspace{-1em}
  
\end{table*}

%% file: tab/daic-item.tex
\begin{table}
    \centering

    \begin{tabular}{l >{\centering\arraybackslash}p{1.3cm} >{\centering\arraybackslash}p{1.3cm} >{\centering\arraybackslash}p{1.3cm}}
        \toprule
        \textbf{Item} & \textbf{CoT} & \textbf{MDAgents} & \textbf{Ours}  \\
        \midrule
        Loss of Interest                      & 37.8 & 45.8 & 38.7  \\
        Depressed Mood                        & 55.8 & 57.1 & 55.7  \\
        Sleep Problems                        & 49.6 & 60.7 & 63.6  \\
        Fat or LowEn                          & 45.7 & 56.6 & 67.0  \\
        Appe or Weight                        & 13.3 & 19.1 & 28.1  \\
        Low Self-Worth                        & 36.8 & 35.2 & 55.2  \\
        Concentration                         & 25.6 & 28.5 & 38.1  \\
        Psychomotor                           & 20.5 & 27.3 & 28.9  \\

        \bottomrule
    \end{tabular}
    \caption{Item-level fine-grained assessment results of F1 score on DAIC-WOZ. `Fat or LowEn', `Appe or Weigh', ` Concentration' and `Psychomotor' refer to `Fatigue or Low Energy', `Appetite or Weight Changes', `Concentration Difficulties' and `Psychomotor Changes'. }
    \label{tab:daic-item}
    \vspace{-1em}
\end{table}  

%% file: tab/ablation.tex
\begin{table}[t]
\centering

\begin{tabular}{@{}ccccc@{}}
\toprule
\multicolumn{2}{c}{\textbf{Module}}          
& \multirow{2}{*}{\textbf{MAE$\downarrow$}} 
& \multirow{2}{*}{\textbf{Kappa$\uparrow$}} 
& \multirow{2}{*}{\textbf{F1-score$\uparrow$}} 
\\

\cmidrule(r){1-2}

\textbf{In-depth} & \textbf{Memory}  &  &   \\ 

\midrule

\ding{51}  & \ding{51}    & \textbf{2.514}  & \textbf{79.8} & \textbf{89.8}
\\
\ding{51}  & \ding{55}    & 3.000 & 58.7 & 79.0   
\\
\ding{55}  & \ding{51}    & 3.314 & 60.4 & 80.1   
\\
\ding{55}  & \ding{55}    & 3.400 & 47.2 & 73.5

\\ 
          
\bottomrule
\end{tabular}
\caption{The results of ablation study.}
\label{tab:ablation}
\vspace{-1em}
\end{table}

%% file: aaai2026.bbl
\begin{thebibliography}{41}
\providecommand{\natexlab}[1]{#1}

\bibitem[{Achiam et~al.(2023)Achiam, Adler, Agarwal, Ahmad, Akkaya, Aleman, Almeida, Altenschmidt, Altman, Anadkat et~al.}]{achiam2023gpt}
Achiam, J.; Adler, S.; Agarwal, S.; Ahmad, L.; Akkaya, I.; Aleman, F.~L.; Almeida, D.; Altenschmidt, J.; Altman, S.; Anadkat, S.; et~al. 2023.
\newblock Gpt-4 technical report.
\newblock \emph{arXiv preprint arXiv:2303.08774}.

\bibitem[{Agarwal, Dias, and Dollfus(2024)}]{agarwal2024analysing}
Agarwal, N.; Dias, G.; and Dollfus, S. 2024.
\newblock Analysing relevance of discourse structure for improved mental health estimation.
\newblock In \emph{Proceedings of the 9th Workshop on Computational Linguistics and Clinical Psychology (CLPsych 2024)}, 127--132.

\bibitem[{Almansoori, Kumar, and Cholakkal(2025)}]{almansoori2025self}
Almansoori, M.; Kumar, K.; and Cholakkal, H. 2025.
\newblock Self-Evolving Multi-Agent Simulations for Realistic Clinical Interactions.
\newblock \emph{arXiv preprint arXiv:2503.22678}.

\bibitem[{Chan et~al.(2024)Chan, Chen, Su, Yu, Xue, Zhang, Fu, and Liu}]{chan2024chateval}
Chan, C.-M.; Chen, W.; Su, Y.; Yu, J.; Xue, W.; Zhang, S.; Fu, J.; and Liu, Z. 2024.
\newblock ChatEval: Towards Better {LLM}-based Evaluators through Multi-Agent Debate.
\newblock In \emph{The Twelfth International Conference on Learning Representations}.

\bibitem[{Chen et~al.(2024{\natexlab{a}})Chen, Su, Zuo, Yang, Yuan, Chan, Yu, Lu, Hung, Qian, Qin, Cong, Xie, Liu, Sun, and Zhou}]{chen2024agentverse}
Chen, W.; Su, Y.; Zuo, J.; Yang, C.; Yuan, C.; Chan, C.-M.; Yu, H.; Lu, Y.; Hung, Y.-H.; Qian, C.; Qin, Y.; Cong, X.; Xie, R.; Liu, Z.; Sun, M.; and Zhou, J. 2024{\natexlab{a}}.
\newblock AgentVerse: Facilitating Multi-Agent Collaboration and Exploring Emergent Behaviors.
\newblock In \emph{The Twelfth International Conference on Learning Representations}.

\bibitem[{Chen et~al.(2024{\natexlab{b}})Chen, Deng, Zhou, Wu, Qian, and Huang}]{chen2024depression}
Chen, Z.; Deng, J.; Zhou, J.; Wu, J.; Qian, T.; and Huang, M. 2024{\natexlab{b}}.
\newblock Depression detection in clinical interviews with LLM-empowered structural element graph.
\newblock In \emph{Proceedings of the 2024 Conference of the North American Chapter of the Association for Computational Linguistics: Human Language Technologies (Volume 1: Long Papers)}, 8181--8194.

\bibitem[{Chen et~al.(2024{\natexlab{c}})Chen, Deng, Zhou, Wu, Qian, and Huang}]{chen-etal-2024-depression}
Chen, Z.; Deng, J.; Zhou, J.; Wu, J.; Qian, T.; and Huang, M. 2024{\natexlab{c}}.
\newblock Depression Detection in Clinical Interviews with {LLM}-Empowered Structural Element Graph.
\newblock In Duh, K.; Gomez, H.; and Bethard, S., eds., \emph{Proceedings of the 2024 Conference of the North American Chapter of the Association for Computational Linguistics: Human Language Technologies (Volume 1: Long Papers)}, 8181--8194. Mexico City, Mexico: Association for Computational Linguistics.

\bibitem[{Chiong et~al.(2021)Chiong, Budhi, Dhakal, and Chiong}]{chiong2021textual}
Chiong, R.; Budhi, G.~S.; Dhakal, S.; and Chiong, F. 2021.
\newblock A textual-based featuring approach for depression detection using machine learning classifiers and social media texts.
\newblock \emph{Computers in Biology and Medicine}, 135: 104499.

\bibitem[{Chowdhery et~al.(2023)Chowdhery, Narang, Devlin, Bosma, Mishra, Roberts, Barham, Chung, Sutton, Gehrmann et~al.}]{chowdhery2023palm}
Chowdhery, A.; Narang, S.; Devlin, J.; Bosma, M.; Mishra, G.; Roberts, A.; Barham, P.; Chung, H.~W.; Sutton, C.; Gehrmann, S.; et~al. 2023.
\newblock Palm: Scaling language modeling with pathways.
\newblock \emph{Journal of Machine Learning Research}, 24(240): 1--113.

\bibitem[{Dorri, Kanhere, and Jurdak(2018)}]{dorri2018multi}
Dorri, A.; Kanhere, S.~S.; and Jurdak, R. 2018.
\newblock Multi-agent systems: A survey.
\newblock \emph{Ieee Access}, 6: 28573--28593.

\bibitem[{Fan et~al.(2025)Fan, Wei, Tang, Chen, Siyuan, Wei, and Huang}]{fan-etal-2025-ai}
Fan, Z.; Wei, L.; Tang, J.; Chen, W.; Siyuan, W.; Wei, Z.; and Huang, F. 2025.
\newblock {AI} Hospital: Benchmarking Large Language Models in a Multi-agent Medical Interaction Simulator.
\newblock In Rambow, O.; Wanner, L.; Apidianaki, M.; Al-Khalifa, H.; Eugenio, B.~D.; and Schockaert, S., eds., \emph{Proceedings of the 31st International Conference on Computational Linguistics}, 10183--10213. Abu Dhabi, UAE: Association for Computational Linguistics.

\bibitem[{Gratch et~al.(2014)Gratch, Artstein, Lucas, Stratou, Scherer, Nazarian, Wood, Boberg, DeVault, Marsella, Traum, Rizzo, and Morency}]{gratch-etal-2014-distress}
Gratch, J.; Artstein, R.; Lucas, G.; Stratou, G.; Scherer, S.; Nazarian, A.; Wood, R.; Boberg, J.; DeVault, D.; Marsella, S.; Traum, D.; Rizzo, S.; and Morency, L.-P. 2014.
\newblock The Distress Analysis Interview Corpus of human and computer interviews.
\newblock In Calzolari, N.; Choukri, K.; Declerck, T.; Loftsson, H.; Maegaard, B.; Mariani, J.; Moreno, A.; Odijk, J.; and Piperidis, S., eds., \emph{Proceedings of the Ninth International Conference on Language Resources and Evaluation ({LREC}'14)}, 3123--3128. Reykjavik, Iceland: European Language Resources Association (ELRA).

\bibitem[{Guo et~al.(2025)Guo, Yang, Zhang, Song, Zhang, Xu, Zhu, Ma, Wang, Bi et~al.}]{guo2025deepseek}
Guo, D.; Yang, D.; Zhang, H.; Song, J.; Zhang, R.; Xu, R.; Zhu, Q.; Ma, S.; Wang, P.; Bi, X.; et~al. 2025.
\newblock Deepseek-r1: Incentivizing reasoning capability in llms via reinforcement learning.
\newblock \emph{arXiv preprint arXiv:2501.12948}.

\bibitem[{Guo et~al.(2024)Guo, Chen, Wang, Chang, Pei, Chawla, Wiest, and Zhang}]{ijcai2024p890}
Guo, T.; Chen, X.; Wang, Y.; Chang, R.; Pei, S.; Chawla, N.~V.; Wiest, O.; and Zhang, X. 2024.
\newblock Large Language Model Based Multi-agents: A Survey of Progress and Challenges.
\newblock In Larson, K., ed., \emph{Proceedings of the Thirty-Third International Joint Conference on Artificial Intelligence, {IJCAI-24}}, 8048--8057. International Joint Conferences on Artificial Intelligence Organization.
\newblock Survey Track.

\bibitem[{Ji et~al.(2022)Ji, Zhang, Ansari, Fu, Tiwari, and Cambria}]{ji-etal-2022-mentalbert}
Ji, S.; Zhang, T.; Ansari, L.; Fu, J.; Tiwari, P.; and Cambria, E. 2022.
\newblock {M}ental{BERT}: Publicly Available Pretrained Language Models for Mental Healthcare.
\newblock In Calzolari, N.; B{\'e}chet, F.; Blache, P.; Choukri, K.; Cieri, C.; Declerck, T.; Goggi, S.; Isahara, H.; Maegaard, B.; Mariani, J.; Mazo, H.; Odijk, J.; and Piperidis, S., eds., \emph{Proceedings of the Thirteenth Language Resources and Evaluation Conference}, 7184--7190. Marseille, France: European Language Resources Association.

\bibitem[{Jiang et~al.(2024)Jiang, Zhang, Cao, Breazeal, Roy, and Kabbara}]{jiang-etal-2024-personallm}
Jiang, H.; Zhang, X.; Cao, X.; Breazeal, C.; Roy, D.; and Kabbara, J. 2024.
\newblock {P}ersona{LLM}: Investigating the Ability of Large Language Models to Express Personality Traits.
\newblock In Duh, K.; Gomez, H.; and Bethard, S., eds., \emph{Findings of the Association for Computational Linguistics: NAACL 2024}, 3605--3627. Mexico City, Mexico: Association for Computational Linguistics.

\bibitem[{Kebe et~al.(2025)Kebe, Girard, Liebenthal, Baker, De~la Torre, and Morency}]{kebe2025llamadrs}
Kebe, G.~Y.; Girard, J.~M.; Liebenthal, E.; Baker, J.; De~la Torre, F.; and Morency, L.-P. 2025.
\newblock LlaMADRS: Prompting Large Language Models for Interview-Based Depression Assessment.
\newblock \emph{arXiv preprint arXiv:2501.03624}.

\bibitem[{Khowaja et~al.(2025)Khowaja, Nkenyereye, Khuwaja, Al~Hamadi, and Dev}]{10681286}
Khowaja, S.~A.; Nkenyereye, L.; Khuwaja, P.; Al~Hamadi, H.; and Dev, K. 2025.
\newblock Depression Detection From Social Media Posts Using Emotion Aware Encoders and Fuzzy Based Contrastive Networks.
\newblock \emph{IEEE Transactions on Fuzzy Systems}, 33(1): 43--53.

\bibitem[{Kim et~al.(2024)Kim, Park, Jeong, Chan, Xu, McDuff, Breazeal, and Park}]{Kim2024MDAgentsAA}
Kim, Y.~H.; Park, C.; Jeong, H.; Chan, Y.~S.; Xu, X.; McDuff, D.; Breazeal, C.; and Park, H.~W. 2024.
\newblock MDAgents: An Adaptive Collaboration of LLMs for Medical Decision-Making.
\newblock In \emph{Neural Information Processing Systems}.

\bibitem[{Kwon et~al.(2023)Kwon, Li, Zhuang, Sheng, Zheng, Yu, Gonzalez, Zhang, and Stoica}]{kwon2023efficient}
Kwon, W.; Li, Z.; Zhuang, S.; Sheng, Y.; Zheng, L.; Yu, C.~H.; Gonzalez, J.; Zhang, H.; and Stoica, I. 2023.
\newblock Efficient memory management for large language model serving with pagedattention.
\newblock In \emph{Proceedings of the 29th Symposium on Operating Systems Principles}, 611--626.

\bibitem[{Li et~al.(2023)Li, Hammoud, Itani, Khizbullin, and Ghanem}]{li2023camel}
Li, G.; Hammoud, H.; Itani, H.; Khizbullin, D.; and Ghanem, B. 2023.
\newblock Camel: Communicative agents for" mind" exploration of large language model society.
\newblock \emph{Advances in Neural Information Processing Systems}, 36: 51991--52008.

\bibitem[{Li et~al.(2024)Li, Zhu, Lin, Li, Jiang, and Zeng}]{li2024zero}
Li, W.; Zhu, Y.; Lin, X.; Li, M.; Jiang, Z.; and Zeng, Z. 2024.
\newblock Zero-shot explainable mental health analysis on social media by incorporating mental scales.
\newblock In \emph{Companion Proceedings of the ACM Web Conference 2024}, 959--962.

\bibitem[{Ohse et~al.(2024)Ohse, Had{\v{z}}i{\'c}, Mohammed, Peperkorn, Fox, Krutzki, Lyko, Mingyu, Zheng, R{\"a}tsch et~al.}]{ohse2024gpt}
Ohse, J.; Had{\v{z}}i{\'c}, B.; Mohammed, P.; Peperkorn, N.; Fox, J.; Krutzki, J.; Lyko, A.; Mingyu, F.; Zheng, X.; R{\"a}tsch, M.; et~al. 2024.
\newblock GPT-4 shows potential for identifying social anxiety from clinical interview data.
\newblock \emph{Scientific Reports}, 14(1): 1--12.

\bibitem[{Qiu et~al.(2024)Qiu, He, Zhang, Li, and Lan}]{qiu-etal-2024-smile}
Qiu, H.; He, H.; Zhang, S.; Li, A.; and Lan, Z. 2024.
\newblock {SMILE}: Single-turn to Multi-turn Inclusive Language Expansion via {C}hat{GPT} for Mental Health Support.
\newblock In Al-Onaizan, Y.; Bansal, M.; and Chen, Y.-N., eds., \emph{Findings of the Association for Computational Linguistics: EMNLP 2024}, 615--636. Miami, Florida, USA: Association for Computational Linguistics.

\bibitem[{Sommers-Flanagan, Zeleke, and Hood(2015)}]{sommers2015clinical}
Sommers-Flanagan, J.; Zeleke, W.~A.; and Hood, M. 2015.
\newblock The clinical interview.
\newblock \emph{The encyclopedia of clinical psychology. John Wiley \& Sons, Inc, Hoboken}, 1--9.

\bibitem[{Team(2024)}]{team2024qwen2}
Team, Q. 2024.
\newblock Qwen2. 5: A party of foundation models.
\newblock \emph{Qwen (Sept. 2024). url: https://qwenlm. github. io/blog/qwen2}, 5.

\bibitem[{Touvron et~al.(2023)Touvron, Lavril, Izacard, Martinet, Lachaux, Lacroix, Rozi{\`e}re, Goyal, Hambro, Azhar, Rodriguez, Joulin, Grave, and Lample}]{Touvron2023LLaMAOA}
Touvron, H.; Lavril, T.; Izacard, G.; Martinet, X.; Lachaux, M.-A.; Lacroix, T.; Rozi{\`e}re, B.; Goyal, N.; Hambro, E.; Azhar, F.; Rodriguez, A.; Joulin, A.; Grave, E.; and Lample, G. 2023.
\newblock LLaMA: Open and Efficient Foundation Language Models.
\newblock \emph{ArXiv}, abs/2302.13971.

\bibitem[{Tran et~al.(2025)Tran, Dao, Nguyen, Pham, O'Sullivan, and Nguyen}]{tran2025multiagentcollaborationmechanismssurvey}
Tran, K.-T.; Dao, D.; Nguyen, M.-D.; Pham, Q.-V.; O'Sullivan, B.; and Nguyen, H.~D. 2025.
\newblock Multi-Agent Collaboration Mechanisms: A Survey of LLMs.
\newblock arXiv:2501.06322.

\bibitem[{Wang et~al.(2024)Wang, Wang, Su, Tong, and Song}]{wang-etal-2024-rethinking-bounds}
Wang, Q.; Wang, Z.; Su, Y.; Tong, H.; and Song, Y. 2024.
\newblock Rethinking the Bounds of {LLM} Reasoning: Are Multi-Agent Discussions the Key?
\newblock In Ku, L.-W.; Martins, A.; and Srikumar, V., eds., \emph{Proceedings of the 62nd Annual Meeting of the Association for Computational Linguistics (Volume 1: Long Papers)}, 6106--6131. Bangkok, Thailand: Association for Computational Linguistics.

\bibitem[{Wang et~al.(2023{\natexlab{a}})Wang, Wei, Schuurmans, Le, Chi, Narang, Chowdhery, and Zhou}]{wang2023selfconsistency}
Wang, X.; Wei, J.; Schuurmans, D.; Le, Q.~V.; Chi, E.~H.; Narang, S.; Chowdhery, A.; and Zhou, D. 2023{\natexlab{a}}.
\newblock Self-Consistency Improves Chain of Thought Reasoning in Language Models.
\newblock In \emph{The Eleventh International Conference on Learning Representations}.

\bibitem[{Wang et~al.(2023{\natexlab{b}})Wang, Xiao, Huang, Yuan, Xu, Guo, Tu, Fei, Leng, Wang et~al.}]{wang2023incharacter}
Wang, X.; Xiao, Y.; Huang, J.-t.; Yuan, S.; Xu, R.; Guo, H.; Tu, Q.; Fei, Y.; Leng, Z.; Wang, W.; et~al. 2023{\natexlab{b}}.
\newblock Incharacter: Evaluating personality fidelity in role-playing agents through psychological interviews.
\newblock \emph{arXiv preprint arXiv:2310.17976}.

\bibitem[{Wei et~al.(2022)Wei, Wang, Schuurmans, Bosma, Xia, Chi, Le, Zhou et~al.}]{wei2022chain}
Wei, J.; Wang, X.; Schuurmans, D.; Bosma, M.; Xia, F.; Chi, E.; Le, Q.~V.; Zhou, D.; et~al. 2022.
\newblock Chain-of-thought prompting elicits reasoning in large language models.
\newblock \emph{Advances in neural information processing systems}, 35: 24824--24837.

\bibitem[{Wu et~al.(2024)Wu, Bansal, Zhang, Wu, Li, Zhu, Jiang, Zhang, Zhang, Liu et~al.}]{wu2024autogen}
Wu, Q.; Bansal, G.; Zhang, J.; Wu, Y.; Li, B.; Zhu, E.; Jiang, L.; Zhang, X.; Zhang, S.; Liu, J.; et~al. 2024.
\newblock AutoGen: Enabling Next-Gen LLM Applications via Multi-Agent Conversation.
\newblock In \emph{ICLR 2024 Workshop on Large Language Model (LLM) Agents}.

\bibitem[{Xiao et~al.(2024)Xiao, Xie, Kuang, Liu, Yang, Peng, Han, and Huang}]{xiao2024healme}
Xiao, M.; Xie, Q.; Kuang, Z.; Liu, Z.; Yang, K.; Peng, M.; Han, W.; and Huang, J. 2024.
\newblock HealMe: Harnessing Cognitive Reframing in Large Language Models for Psychotherapy.
\newblock In \emph{Proceedings of the 62nd Annual Meeting of the Association for Computational Linguistics (Volume 1: Long Papers)}, 1707--1725.

\bibitem[{Xu et~al.(2024)Xu, Yao, Dong, Gabriel, Yu, Hendler, Ghassemi, Dey, and Wang}]{xu2024mental}
Xu, X.; Yao, B.; Dong, Y.; Gabriel, S.; Yu, H.; Hendler, J.; Ghassemi, M.; Dey, A.~K.; and Wang, D. 2024.
\newblock Mental-llm: Leveraging large language models for mental health prediction via online text data.
\newblock \emph{Proceedings of the ACM on Interactive, Mobile, Wearable and Ubiquitous Technologies}, 8(1): 1--32.

\bibitem[{Xu and Jiang(2024)}]{xu-jiang-2024-multi}
Xu, Z.; and Jiang, J. 2024.
\newblock Multi-dimensional Evaluation of Empathetic Dialogue Responses.
\newblock In Al-Onaizan, Y.; Bansal, M.; and Chen, Y.-N., eds., \emph{Findings of the Association for Computational Linguistics: EMNLP 2024}, 2066--2087. Miami, Florida, USA: Association for Computational Linguistics.

\bibitem[{Yang et~al.(2024{\natexlab{a}})Yang, Yang, Zhang, Hui, Zheng, Yu, Li, Liu, Huang, Wei et~al.}]{yang2024qwen2}
Yang, A.; Yang, B.; Zhang, B.; Hui, B.; Zheng, B.; Yu, B.; Li, C.; Liu, D.; Huang, F.; Wei, H.; et~al. 2024{\natexlab{a}}.
\newblock Qwen2. 5 technical report.
\newblock \emph{arXiv preprint arXiv:2412.15115}.

\bibitem[{Yang et~al.(2024{\natexlab{b}})Yang, Zhang, Kuang, Xie, Huang, and Ananiadou}]{yang2024mentallama}
Yang, K.; Zhang, T.; Kuang, Z.; Xie, Q.; Huang, J.; and Ananiadou, S. 2024{\natexlab{b}}.
\newblock MentaLLaMA: interpretable mental health analysis on social media with large language models.
\newblock In \emph{Proceedings of the ACM Web Conference 2024}, 4489--4500.

\bibitem[{Yang et~al.(2024{\natexlab{c}})Yang, Wang, Chen, Wang, Pu, Gao, Huang, Song, and Huang}]{yang-etal-2024-psychogat}
Yang, Q.; Wang, Z.; Chen, H.; Wang, S.; Pu, Y.; Gao, X.; Huang, W.; Song, S.; and Huang, G. 2024{\natexlab{c}}.
\newblock {P}sycho{GAT}: A Novel Psychological Measurement Paradigm through Interactive Fiction Games with {LLM} Agents.
\newblock In Ku, L.-W.; Martins, A.; and Srikumar, V., eds., \emph{Proceedings of the 62nd Annual Meeting of the Association for Computational Linguistics (Volume 1: Long Papers)}, 14470--14505. Bangkok, Thailand: Association for Computational Linguistics.

\bibitem[{Yu et~al.(2025)Yu, Yu, Wei, Zhang, and Qian}]{yu-etal-2025-beyond}
Yu, Y.; Yu, R.; Wei, H.; Zhang, Z.; and Qian, Q. 2025.
\newblock Beyond Dialogue: A Profile-Dialogue Alignment Framework Towards General Role-Playing Language Model.
\newblock In Che, W.; Nabende, J.; Shutova, E.; and Pilehvar, M.~T., eds., \emph{Proceedings of the 63rd Annual Meeting of the Association for Computational Linguistics (Volume 1: Long Papers)}, 11992--12022. Vienna, Austria: Association for Computational Linguistics.
\newblock ISBN 979-8-89176-251-0.

\bibitem[{Zhao et~al.(2024)Zhao, Li, Lu, Weber, Lee, Chu, and Wermter}]{zhao-etal-2024-enhancing-zero}
Zhao, X.; Li, M.; Lu, W.; Weber, C.; Lee, J.~H.; Chu, K.; and Wermter, S. 2024.
\newblock Enhancing Zero-Shot Chain-of-Thought Reasoning in Large Language Models through Logic.
\newblock In Calzolari, N.; Kan, M.-Y.; Hoste, V.; Lenci, A.; Sakti, S.; and Xue, N., eds., \emph{Proceedings of the 2024 Joint International Conference on Computational Linguistics, Language Resources and Evaluation (LREC-COLING 2024)}, 6144--6166. Torino, Italia: ELRA and ICCL.

\end{thebibliography}
